\pdfoutput=1

\documentclass[11pt]{article}

\usepackage{EMNLP2023}

\usepackage{times}
\usepackage{latexsym}

\usepackage[T1]{fontenc}

\usepackage[utf8]{inputenc}

\usepackage{microtype}

\usepackage{inconsolata}
\usepackage{booktabs} 
\usepackage{multirow} 
\usepackage{array} 
\usepackage{graphicx}
\usepackage{float}

\title{Cure or Poison? Embedding Instructions Visually Alters Hallucination in Vision-Language Models}

\author{
Zhaochen Wang\textsuperscript{1} \quad
Yiwei Wang\textsuperscript{2} \quad
Yujun Cai\textsuperscript{1} \\
\textsuperscript{1}The University of Queensland \quad
\textsuperscript{2}University of California, Merced \\
\texttt{zhaochen.wang@uqconnect.edu.au}
}

\begin{document}
\maketitle
\begin{abstract}
Vision-Language Models (VLMs) often suffer from hallucination, partly due to challenges in aligning multimodal information. We propose Prompt-in-Image, a simple method that embeds textual instructions directly into images. This removes the need for separate text inputs and forces the model to process all content through the visual channel. We evaluate this method on three popular open-source VLMs: Qwen2.5-VL, LLaVA-1.5, and InstructBLIP. The results reveal sharp differences. Prompt-in-Image improves Qwen2.5-VL’s performance, increasing POPE accuracy by 4.1\% (from 80.2\% to 84.3\%) and also reducing hallucination rates on MS-COCO. In contrast, LLaVA-1.5 and InstructBLIP experience a severe performance drop, with accuracy falling from around 84\% to near-random levels. Through detailed analysis, we found that CLIP-based encoders in LLaVA and InstructBLIP exhibit excessive attention bias toward embedded text regions, disrupting visual understanding. In contrast, Qwen's vision encoder handles text-embedded images robustly. Crucially, Prompt-in-Image reduces Qwen's modality gap, enhancing cross-modal alignment by unifying information processing through a single modality.

\end{abstract}

\section{Introduction}
Most modern Vision-Language Models (VLMs) follow a standard architecture: a visual encoder (typically ViT), a projector, and a language model (LLM decoder). Since visual and textual components are pre-trained separately~\cite{rabinovich2023predicting}, this approach introduces inherent cross-modal alignment challenges, significantly hampering the model's overall performance. One prominent manifestation of these alignment issues is the language bias phenomenon, where VLMs disproportionately rely on textual information while ignoring visual information~\cite{niu2021counterfactual, wang2024mdpo, wang2025text}. To address these cross-modal alignment challenges, previous approaches have focused on improving cross-modal fusion. However, instead of enhancing cross-modal integration, we ask whether we can avoid cross-modal alignment challenges entirely by relying solely on single-modality information.

\begin{figure}
    \centering
    \includegraphics[width=1\linewidth]{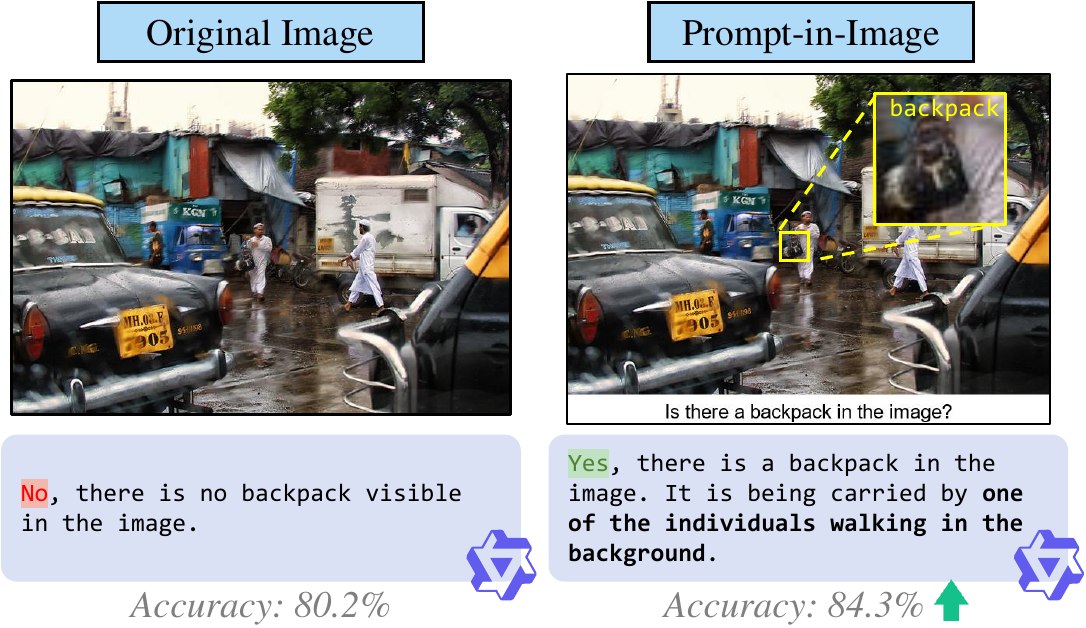}
    \caption{An example of Prompt-in-Image: the text instruction is directly embedded into the image. Using Prompt-in-Image, Qwen2.5-vl performance improves.}
    \label{fig:visual_prompting_sample}
\end{figure}

We propose Prompt-in-Image (Figure~\ref{fig:visual_prompting_sample}), which embeds textual instructions directly into images. By forcing models to process all information through the visual channel, this approach may enhance fusion and reduce alignment issues.

We use hallucination as our primary evaluation task for two key reasons: (1) hallucination represents a major challenge in VLM development, where models describe non-existent objects or miss key visual details, seriously affecting VLM performance and reliability; and (2) hallucination is highly correlated with modality alignment issues~\cite{liu2024survey}, making it an ideal testbed for evaluating our approach.

We use POPE, a representative hallucination benchmark, to extensively test three popular VLMs: Qwen2.5-VL, InstructBLIP and LLaVA-1.5. Surprisingly, we observe opposite effects. On POPE, Qwen2.5-VL's accuracy improves by 4–5\%, while LLaVA-1.5's performance drops dramatically from 84\% to 55\% (a near-complete collapse). Similarly, InstructBLIP shows consistent behavior with LLaVA-1.5, declining from 74.4\% to 54\%. Our analysis reveals two key findings. First, LLaVA and InstructBLIP's collapse comes from its CLIP-based encoder's strong text bias, which gives too much attention to embedded text regions. This creates severe hallucinations. In contrast, Qwen's vision encoder handles text-embedded images much better. Second, Prompt-in-Image effectively reduces Qwen's modality gap, improving cross-modal alignment and boosting performance.

In summary, our contributions are threefold:
\begin{enumerate}
    \item We propose \textit{Prompt-in-Image}, a novel input strategy that embeds text into images to improve modality integration.
    \item We conduct systematic evaluations on Qwen-VL, InstructBLIP and LLaVA-1.5, revealing divergent effects of Prompt-in-Image on hallucination performance.
    \item We conduct an in-depth analysis of the performance gap and explain  how Prompt-in-Image improves performance by reducing modality gaps and enhancing alignment
\end{enumerate}

The rest of this paper is organized as follows. Section~\ref{sec:related_works} reviews related work on VLM architectures, hallucination problems, and existing mitigation methods. Section~\ref{sec:method} presents our Prompt-in-Image method, including the design details, evaluation benchmarks (POPE and MS-COCO), tested models, and experimental configurations. Section~\ref{sec:results} reports our experimental results, showing Prompt-in-Image's contrasting effects on different models. Section~\ref{sec:analysis} provides an in-depth analysis to explain these divergent outcomes. Finally, Section~\ref{sec:conclusion} concludes the paper and discusses future directions. 

\section{Related Works}
\label{sec:related_works}
\subsection{Vision-Language Models and Hallucination}
Vision-Language Models (VLMs) have rapidly evolved in recent years, with many powerful models like GPT-4o~\cite{openai2024gpt4o}, LLaVA~\cite{liu2023visual}, and Qwen-VL~\cite{bai2025qwen2} achieving impressive performance. These models typically share similar architectures: a vision encoder to process images, a projection layer, and a language model to generate text. Despite their success in various tasks, hallucination remains a critical challenge for VLMs. Hallucinations can be categorized into two types. Judgement hallucination occurs when the model's response to a user's query is in disagreement with the actual visual data. Description hallucination is a failure to faithfully depict the visual information~\cite{liu2024survey}. To comprehensively evaluate and quantify this problem, researchers have proposed various VLM hallucination evaluation methods and benchmarks, including POPE~\cite{li2023evaluating}, NOPE~\cite{lovenia2023negative}, CHAIR~\cite{rohrbach2018object}, MMHal-Bench~\cite{sun2023aligning}, and AMBER~\cite{wang2023amber}.

\subsection{Hallucination Causes and Mitigation Methods}
The causes of hallucination in VLMs are complex and multifaceted. Several important factors contribute to this problem, such as data bias~\cite{liu2023mitigating}, limitations in vision encoders~\cite{li2024monkey,cho2022fine,gong2024damro}, poor modality alignment~\cite{sun2023aligning}, and the inherent hallucination of LLMs.
To address these issues, various methods have been proposed. Among them, training-free contrastive decoding (CD) strategies have shown effectiveness in reducing hallucination. Contrastive decoding reduces hallucination by comparing model outputs from original and perturbed inputs—such as visually noised images or modified instructions—to suppress over-reliance on language priors. Representative examples include Visual Contrastive Decoding (VCD)~\cite{leng2024mitigating} and Instruction Contrastive Decoding (ICD)~\cite{wang2024mitigating}. However, these methods also come with limitations, including slower inference speed and limited performance gains. Moreover, some recent studies~\cite{yin2025mirage} have argued that such decoding strategies may be entirely unrelated to the original objective of hallucination mitigation.

\begin{figure*}
    \centering
    \includegraphics[width=1\linewidth]{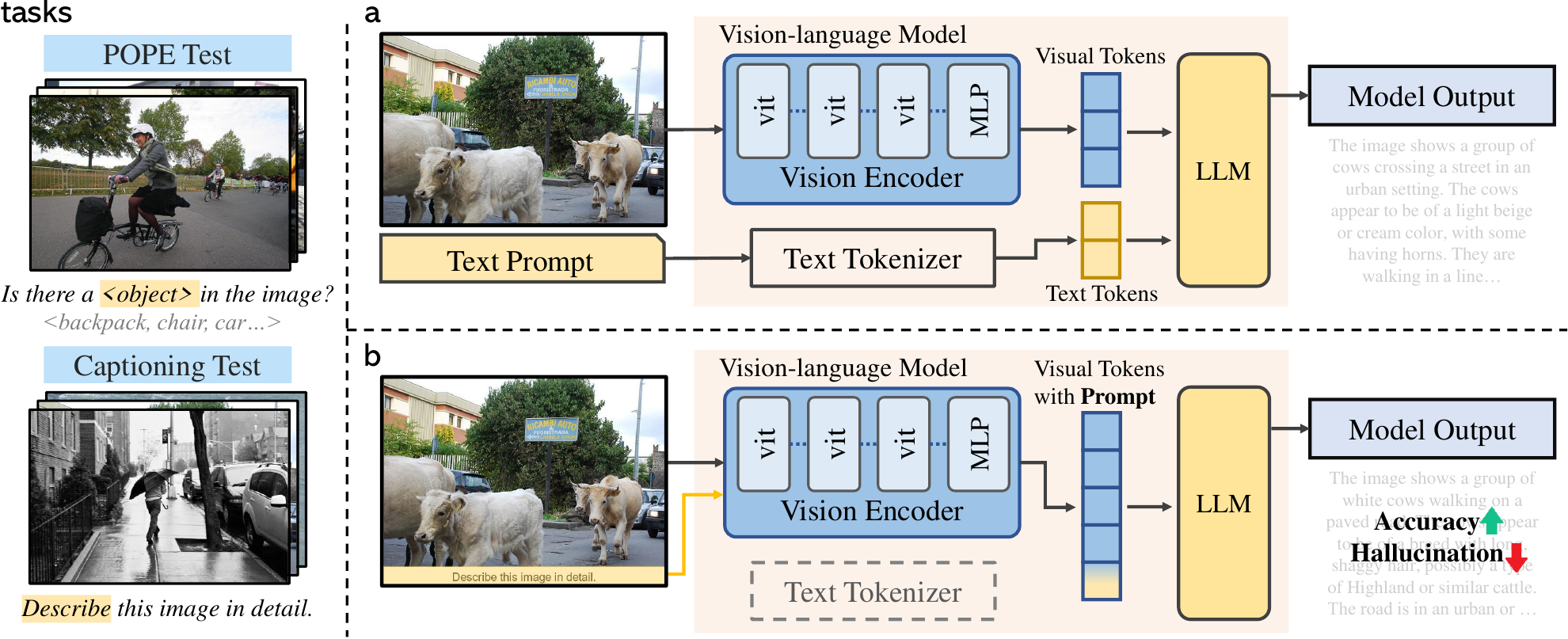}
    \caption{Comparison of interaction paradigms. (a) Traditional VLM interaction requires both an image and a separate text prompt as input. (b) Prompt-in-Image embeds the instruction directly into the image, allowing users to provide only the image without any extra textual input.}
    \label{fig:overview}
\end{figure*}

\section{Method}
\label{sec:method}

\subsection{Prompt-in-Image Design}
In traditional VQA tasks, users provide both an image and textual instructions (prompt), and VLMs process both visual and textual inputs to complete the task. To enable models to rely on single-modality information, we adopt a straightforward approach: directly embedding instructions into the image, similar to movie subtitles. We then provide only the image to the VLM, eliminating separate textual input. We call this approach "Prompt-in-Image."

To create Prompt-in-Image, we render the question text at the bottom of each image. To avoid occluding image content, we add a separate white rectangular area below each image. The question is rendered in black Arial font (26pt), ensuring machine readability. This text region covers only about 5\% of the total image height, minimizing interference with the original visual content. To control for visual changes unrelated to text, we include a white blank box condition: a control group where the same white rectangular area is added without any text. The image size in this group is kept identical to that in the Prompt-in-Image condition.

\subsection{Benchmarks and Evaluation Metrics}
We evaluate Prompt-in-Image on two complementary benchmarks:

\textbf{POPE (Polling-based Object Probing Evaluation)}~\cite{li2023evaluating}: This dataset evaluates object existence detection using binary questions such as “\textit{Is there a <object> in the image?}”, where \textit{<object>} is selected from three splits: random (randomly selected objects), popular (frequently occurring objects), and adversarial (objects closely related to those in the image). We focus on the adversarial subset, which is the most challenging and closely reflects real-world hallucination scenarios. The complete adversarial dataset consists of 3,000 questions (500 images × 6 questions per image). To balance comprehensive evaluation with computational efficiency, we randomly selected 1,000 questions for testing. Model performance is measured using accuracy and F1 score.

\textbf{MS-COCO Caption}: We randomly select 500 images from the MS-COCO 2017~\cite{lin2014microsoft} validation set and prompt the VLMs with ``\textit{Describe this image in detail.}'' Hallucination is evaluated using the \textbf{CHAIR} metric~\cite{rohrbach2018object}, which compares generated captions against ground-truth object labels. We report two scores: \textbf{CHAIRi}, the proportion of hallucinated objects among all mentioned objects; and \textbf{CHAIRs}, the proportion of captions containing at least one hallucinated object.

These two datasets are well-established benchmarks and provide a comprehensive evaluation framework, covering both binary question answering and open-ended generation tasks.

\section{Experimental Results}
\label{sec:results}

\begin{table*}
\centering
\footnotesize
\begin{tabular}{c|ccc|ccc|ccc}
\toprule
\multirow{2}{*}{\textbf{Setting}} & \multicolumn{3}{c|}{\textbf{Qwen2.5-VL-7B}} & \multicolumn{3}{c|}{\textbf{InstructBLIP-7B}} & \multicolumn{3}{c}{\textbf{LLaVA-v1.5-7B}} \\
\cmidrule{2-10}
                 & Acc. & F1 & Yes Ratio & Acc. & F1 & Yes Ratio& Acc. & F1 & Yes Ratio\\
\midrule
\rowcolor[gray]{0.9}
Baseline         & 80.2 & 0.76 & 0.32 & 74.4 & 0.72 & 0.53 & 84.0 & 0.86 & 0.62 \\
Control          & 80.5{\footnotesize\color{green}(+0.3)} & 0.77 & 0.33 & 75.0{\footnotesize\color{green}(+0.6)} & 0.77 & 0.53 & 84.0{\footnotesize\color{gray}(±0)} & 0.86 & 0.62 \\
Hybrid           & 83.0{\footnotesize\color{green}(+2.8)} & 0.81 & 0.38 & 63.2{\footnotesize\color{red}(-11.2)} & 0.69 & 0.68 & 64.0{\footnotesize\color{red}(-20.0)} & 0.74 & 0.82 \\
Prompt-in-Image & \textbf{84.3}{\footnotesize\color{green}(+4.1)} & \textbf{0.82} & 0.38 & -- & -- & -- & -- & -- & -- \\
Prompt-in-Image$^\dagger$ & 82.1{\footnotesize\color{green}(+1.9)} & 0.81 & 0.43 & 54.0{\footnotesize\color{red}(-20.4)} & 0.70 & 0.99 & 55.0{\footnotesize\color{red}(-29.0)} & 0.70 & 0.99 \\
\bottomrule
\end{tabular}
\caption{\label{tab:vlm-comparison}
Performance of different input settings on three VLMs. Prompt-in-Image (without instruction) is only applicable to Qwen2.5-VL. $^\dagger$With explicit instruction "\textit{Answer the questions in the image}".
}
\end{table*}

\begin{table}
\centering
\begin{tabular}{c|cc}
\toprule
\multirow{2}{*}{\textbf{Setting}} & \textbf{CHAIRs} & \textbf{CHAIRi} \\
                                  & \textbf{(\%)} & \textbf{(\%)} \\
\midrule
\rowcolor[gray]{0.9}
Baseline         & 32.3 & 8.8 \\
Hybrid           & 34.2{\footnotesize\color{red}(+1.9)} & 9.9{\footnotesize\color{red}(+1.1)} \\
Prompt-in-Image & 24.7{\footnotesize\color{green}(-7.6)} & 6.7{\footnotesize\color{green}(-2.1)} \\
\bottomrule
\end{tabular}
\caption{\label{tab:chairs}
Hallucination rates on MS-COCO using CHAIR metrics. Prompt-in-Image reduces both CHAIRs and CHAIRi compared to other input modes.
}
\end{table}
\label{sec:coco}

\subsection{Models}
We evaluate our method on two widely used open-source VLMs: \textbf{Qwen2.5-VL-7B}~\cite{bai2025qwen2}, \textbf{InstructBLIP-vicuna-7B}~\cite{dai2023instructblipgeneralpurposevisionlanguagemodels} and \textbf{LLaVA-v1.5-7B}~\cite{liu2023visual}. All three models follow similar transformer-based architectures and demonstrate strong performance across multimodal tasks. Importantly, all three are capable of accurately recognizing questions embedded in images, making them suitable for evaluating the effectiveness of Prompt-in-Image.

\subsection{Experimental Configuration}
We evaluate four input configurations to test the effect of Prompt-in-Image:

\begin{itemize}
    \item \textbf{Baseline:} Original image + text prompt.
    \item \textbf{Prompt-in-Image:} Image with embedded question + no text prompt (system messages like ``You are a helpful assistant.'' may still be present).
    \item \textbf{Hybrid:} Image with embedded question + text prompt.
    \item \textbf{Control:} Image with a blank white box (no text) + text prompt. This controls for the visual change introduced by the prompt region.
\end{itemize}

All experiments are conducted with identical inference parameters and fixed random seeds to ensure reproducibility. The decoding temperature is set to 0.7 (default). These configurations allow us to isolate the effects of visual input, textual input, and their interaction.

VLMs behave very differently when processing our Prompt-in-Image samples. Qwen2.5-VL works quite naturally with embedded questions—it can directly read and answer questions that are placed in the image without needing any extra instructions from us. LLaVA-1.5 and InstructBLIP, however, act differently. When we don't give them specific text instructions, they tend to treat embedded text as just another part of the image to describe. Instead of answering the embedded question, they often give general descriptions of what they see.

To ensure fair comparison, we give InstructBLIP and LLaVA-1.5 a clear instruction for all Prompt-in-Image tests: "\textit{Answer the questions in the image.}" This prevents LLaVA from just describing the image and ensures both models are actually trying to answer the embedded questions. Additionally, for experimental rigor, we conducted the same experiments with the Qwen model (see Table~\ref{tab:vlm-comparison}). We will analyze these results in detail in the next section.

\begin{figure*}[!h]
    \centering
    \includegraphics[width=1\linewidth]{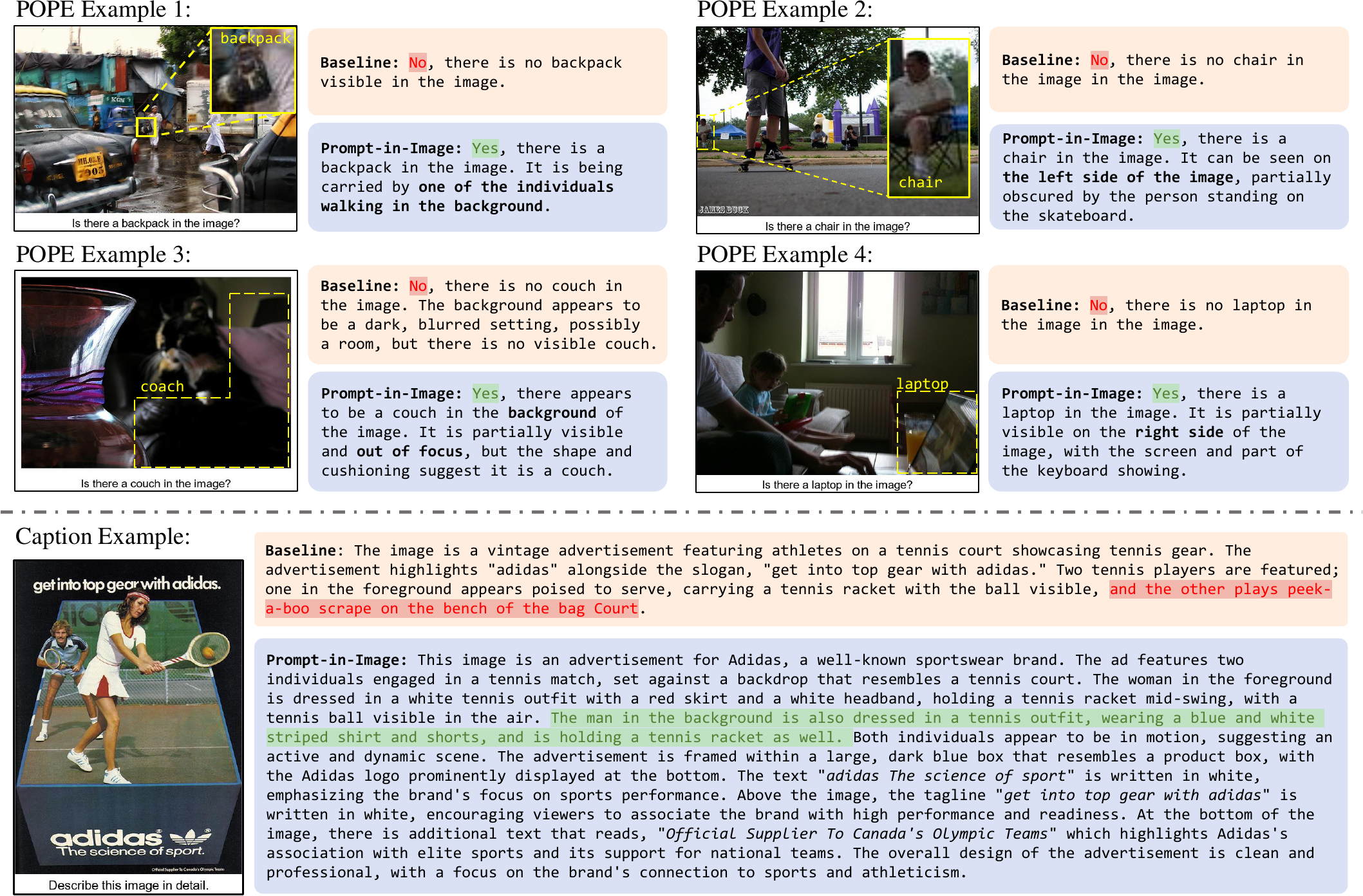}
    \caption{Case studies on Qwen2.5-VL. Top: POPE examples comparing baseline  and Prompt-in-Image responses for object presence detection. Bottom: An MS-COCO captioning example. Prompt-in-Image helps the model generate more detailed answers and with lower hallucination.}
    \label{fig:qwen_case_study}
\end{figure*}

\subsection{POPE Evaluation}
Table~\ref{tab:vlm-comparison} summarizes the performance comparison between three VLMs on POPE. The results reveal striking differences between models:

\textbf{Qwen2.5-VL} demonstrates consistent improvements with Prompt-in-Image. Accuracy increases by 4.1\% (80.2\% → 84.3\%). The Hybrid configuration also shows gains (+2.8\%), though slightly lower than Prompt-in-Image. We examine specific examples to better understand how Prompt-in-Image changes Qwen's behavior (Figure~\ref{fig:qwen_case_study}, Top). The results show clear differences between baseline and Prompt-in-Image responses. Surprisingly, we find that Prompt-in-Image helps the model detect small, unusual, or partially hidden objects that it missed before. More importantly, Qwen doesn't just identify these challenging objects—it can also tell us exactly where they are located in the image (e.g., "on the right side of the image"). This suggests that Prompt-in-Image actually improves Qwen's visual understanding and ability to ground objects in the scene, rather than simply making it say "yes" more often to questions.

\textbf{LLaVA-v1.5} and \textbf{InstructBLIP} exhibit catastrophic performance degradation. LLaVA-v1.5 drops dramatically from 84.0\% baseline accuracy to 55.0\%, while InstructBLIP shows very similar behavior, declining from 74.4\% to 54.0\%. More concerning, both models show Yes Ratios jumping to 0.99, indicating they default to "yes" for nearly all queries. This represents a complete loss of discriminative capability for both models.

In all models, the control groups perform nearly identically to the baseline, confirming that the performance changes are caused by the presence of text in the image, rather than layout or formatting changes.

\subsection{MS-COCO Validation}
Given Qwen2.5-VL's promising results on POPE, we conducted further evaluation using MS-COCO Caption to assess performance on open-ended generation tasks. We tested 500 images, and evaluated model outputs using the CHAIR metric. Prompt-in-Image yields consistent improvements across both hallucination metrics. In Table~\ref{tab:chairs}, CHAIRs decreases by 7.6\% (32.3\% → 24.7\%) and CHAIRi by 2.1\% (8.8\% → 6.7\%), indicating reduced hallucination at both sentence and instance levels. Figure~\ref{fig:qwen_case_study} (Bottom) shows a good example of this improvement. With Prompt-in-Image, the model generates much more detailed captions, including specific details like clothing colors and small text visible in the image. This is interesting because the model actually says more and gives more details, but makes fewer mistakes. Prompt-in-Image helps the model see and describe what's really there, rather than just making up information.

\section{Divergent Effects of Prompt-in-Image}
\label{sec:analysis}

Our experiments reveal contradictory effects of Prompt-in-Image across different models. This section addresses two critical questions: 

\begin{itemize}
    \item Why does InstructBLIP and LLaVA's performance degrade with Prompt-in-Image?
    \item How does Prompt-in-Image enhance Qwen's performance?
\end{itemize}
We examine both questions in detail in this section.

\subsection{Why Prompt-in-Image Hurts LLaVA}

\begin{figure*}
    \centering
    \includegraphics[width=1\linewidth]{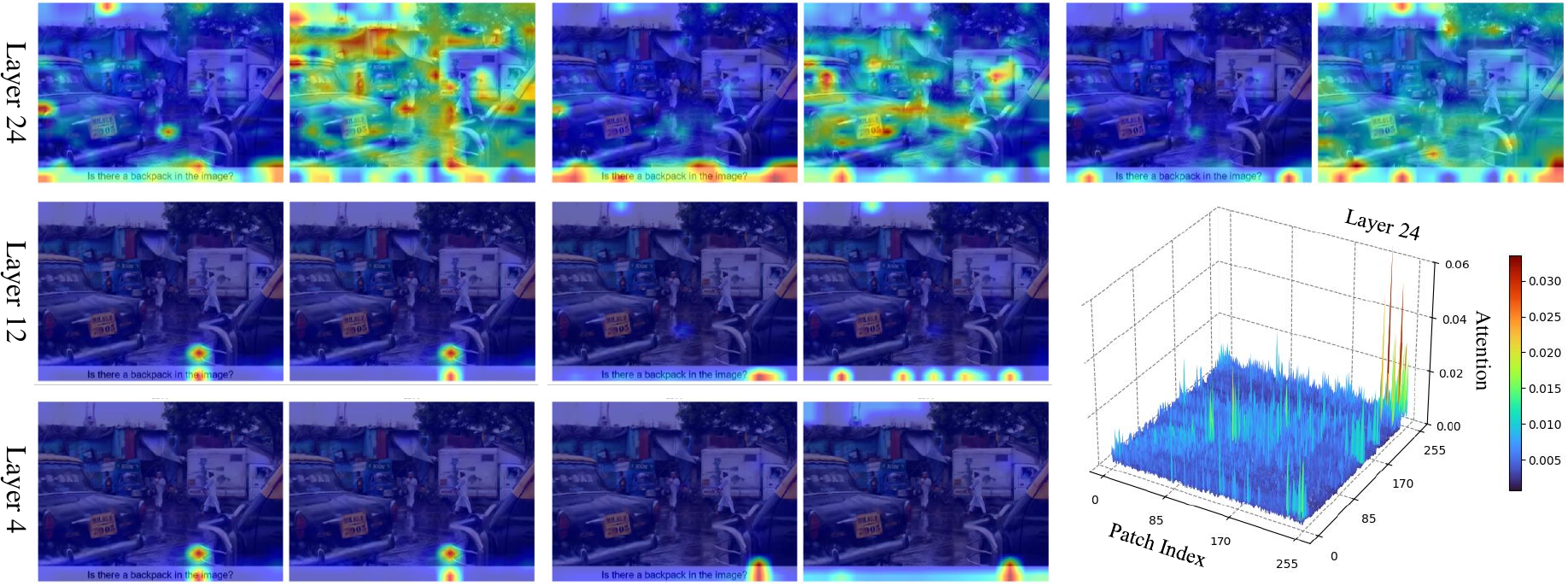}
    \caption{CLIP attention visualization across layers 4, 12, and 24 on the example image. Each row shows patch-level attention weights for Prompt-in-Image (with embedded text) versus Control Group (text-free images). Deep layers (layer 24) exhibit strong attention bias toward text regions.}
    \label{fig:clip_attention}
\end{figure*}

\begin{figure*}[!h]
    \centering
    \includegraphics[width=1\linewidth]{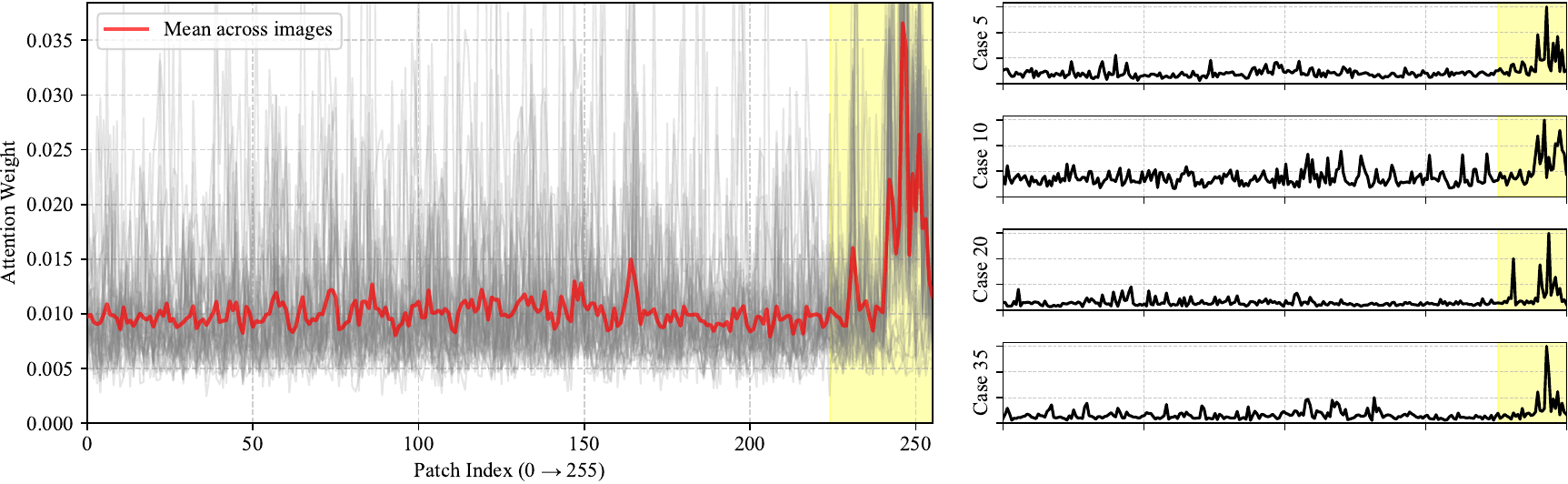}
    \caption{Self-attention analysis of CLIP's final layer across 35 Prompt-in-Image images. The \colorbox{pink}{red line} shows average attention weights across all images, while \colorbox{gray!30}{gray lines} represent individual samples. The \colorbox{yellow!30}{yellow region} corresponds to text-related patches, which consistently receive much higher self-attention scores.}
    \label{fig:attention_lines}
\end{figure*}

We analyze CLIP ViT-L/14~\cite{radford2021learning}, which serves as the visual encoder in LLaVA-v1.5 and is also closely related to the visual encoder in InstructBLIP. We visualize the patch-level attention weights across different layers (Figure~\ref{fig:clip_attention}), using the average attention across all heads. Specifically, we examine three representative layers: a shallow layer (layer 4), a middle layer (layer 12), and a deep layer (layer 24). The goal is to compare how CLIP processes two types of input: images with embedded text (Prompt-in-Image) and their text-free counterparts (Control Group).

We observe that the attention distributions in shallow (layer 4) and middle layers (layer 12) are similar across both image types, suggesting limited sensitivity to embedded text at early stages. However, in the deep layer (layer 24), the difference becomes more pronounced: CLIP shows a clear tendency to focus attention heavily on the text region. 

\begin{figure}[!h]
    \centering
    \includegraphics[width=1\linewidth]{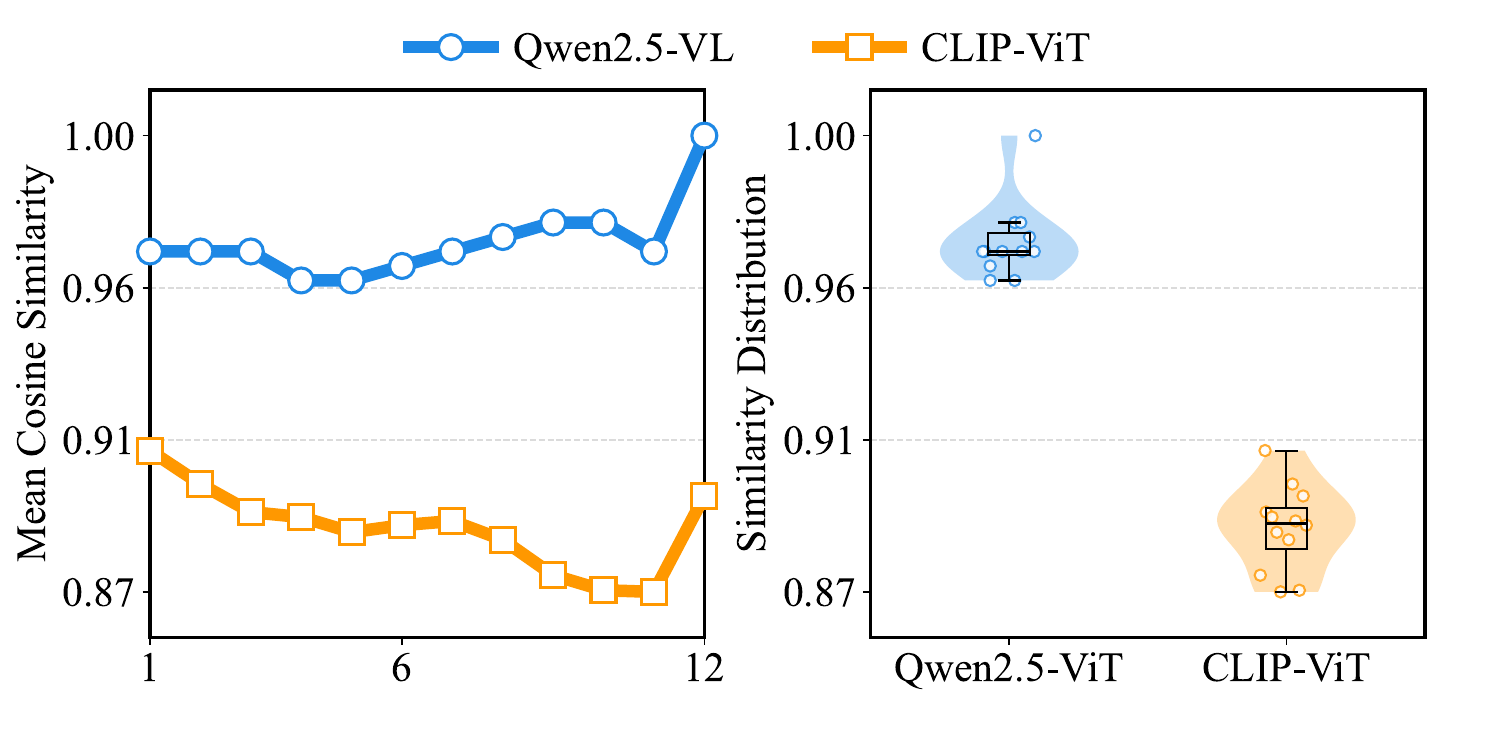}
    \caption{Layer-wise similarity comparison between Qwen2.5‑ViT and CLIP on an example image pair.}
    \label{fig:Qwen_vs_clip}
\end{figure}

\begin{figure*}[!ht]
    \centering
    \includegraphics[width=1\linewidth]{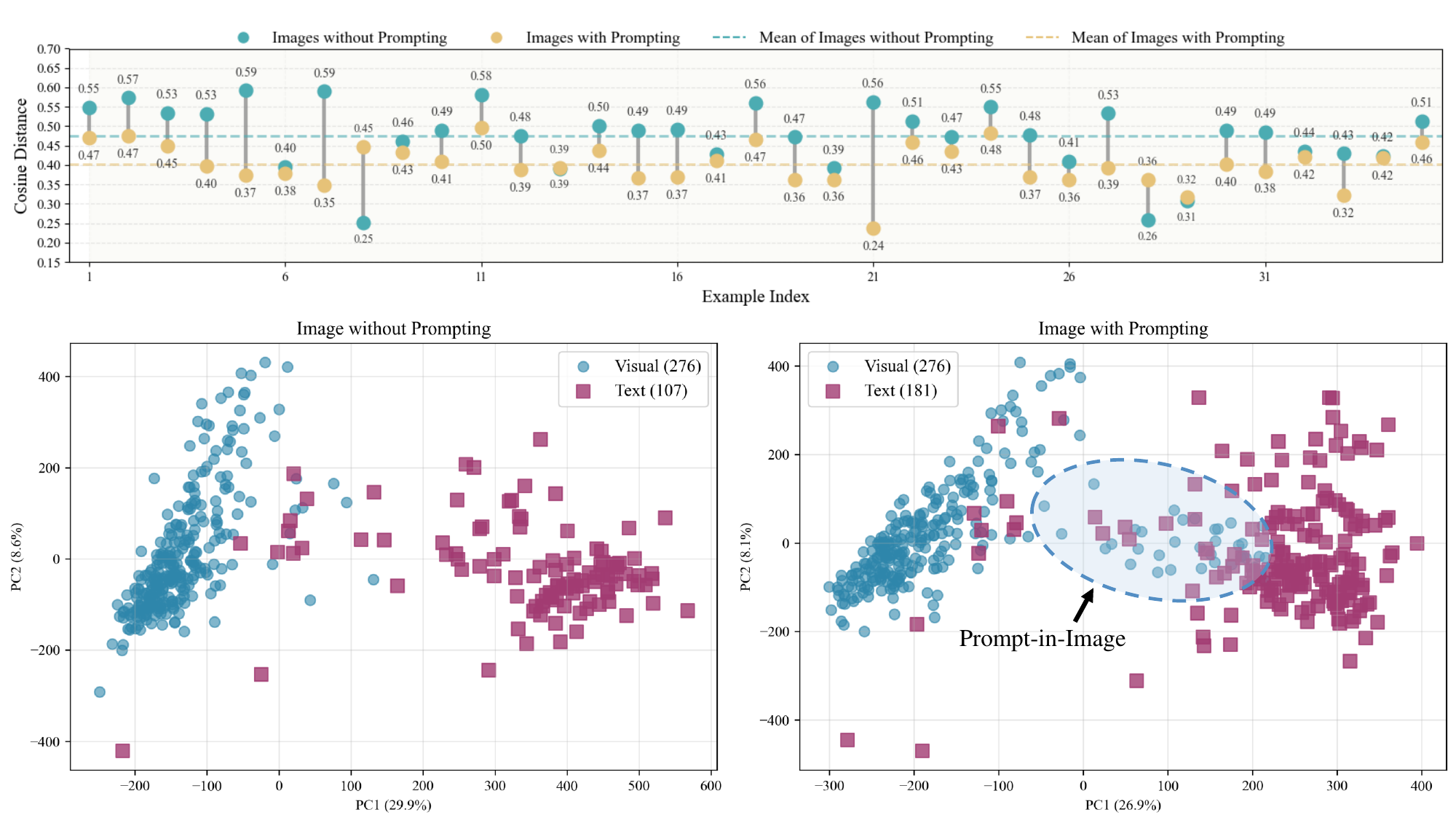}
    \caption{ Modality gap analysis for Qwen2.5-VL. Top: Comparison of average cosine distances between image and caption embeddings across 35 samples. Most Prompt-in-Image samples (yellow) show lower modality gaps compared to baseline (blue). Bottom: PCA visualization of image (blue) and text (purple) embeddings for one example. Prompt-in-Image (right) brings the two modalities closer together compared to the baseline condition (left).}
    \label{fig:token_embedding}
\end{figure*}

To further confirm CLIP's text bias, we quantify this effect by analyzing 35 randomly selected Prompt-in-Image images from the POPE dataset. We examine the self-attention patterns in the final layer (layer 24) of the CLIP encoder, focusing on the diagonal of the attention matrix, which shows how much attention each patch gives to itself. 
Figure~\ref{fig:attention_lines} shows that the last 32 patches, corresponding to the text region, consistently have very high self-attention scores. This confirms that CLIP is overly sensitive to embedded text, giving too much attention to text regions in deeper layers.Many previous works \cite{darcet2023vision, gong2024damro, zhang2024dhcp} suggest that excessive attention to certain visual tokens can lead to hallucinations in VLM outputs. When attention weights concentrate disproportionately on specific visual tokens (like those containing embedded text) the model loses its ability to balance local and global visual information. These dominant tokens can override both visual context and linguistic priors, causing the model to ignore actual image content and default to affirmative responses regardless of other visual evidence.

Additionally, we extract features from all transformer layers when processing Prompt-in-Image and Control image pairs, computing cosine similarity across patches at each layer. Figure~\ref{fig:Qwen_vs_clip} shows the similarity profiles for the final 12 layers, where high-level semantic representations emerge.

The results reveal a striking difference: Qwen-ViT maintains consistently high similarity (>0.95) between Prompt-in-Image and Control images throughout its deep layers, while CLIP-ViT shows declining similarity in the same layers. This indicates that Qwen's vision encoder preserves image semantics despite embedded text, whereas CLIP becomes increasingly sensitive to textual elements in deeper layers.

This robustness likely stems from Qwen's diverse pretraining regime, which includes not only standard image-caption pairs but also interleaved image-text documents and OCR data \cite{bai2025qwen2}. By learning to process images with naturally embedded text during pretraining, Qwen-ViT develops representations that treat text as a normal visual element rather than a disruptive signal—explaining why it can successfully interpret visual prompts without catastrophic attention shifts.

\subsection{Why Prompt-in-Image Helps Qwen}

In contrast to InstructBLIP and LLaVA-1.5, Qwen2.5-VL shows consistent performance gains under Prompt-in-Image. On the POPE dataset, accuracy improves by 4–5\%, and on the open-ended caption generation task for MS-COCO images, hallucination rates also decrease.

We propose a possible explanation: Prompt-in-Image unifies the input modality through the visual channel, thereby enhancing modality fusion and mitigating alignment issues. This hypothesis is supported by two observations.

\textbf{Text input disrupts Prompt-in-Image performance}: As shown in Section~\ref{sec:results}, the Hybrid setting (image with embedded question and separate text prompt) performs worse than the Prompt-in-Image setting (image with embedded question only). This suggests that introducing an additional modality (i.e., text) does not help the model and instead degrades performance. In contrast, Prompt-in-Image consolidates all relevant information into a single modality, which helps the model focus and enhances overall performance.

\textbf{Prompt-in-Image reduces the modality gap}: Previous studies~\cite{liang2022mind} have identified the "modality gap" phenomenon in vision-language models. While these models are designed to map images and text into a shared representation space, different modalities actually end up clearly separated in this space. This separation negatively affects performance across multiple downstream tasks, and reducing this gap has been shown to improve model performance~\cite{role2025fill}.

Ideally, image embeddings and their corresponding caption embeddings should overlap closely in the representation space, indicating good cross-modal alignment. To test whether Prompt-in-Image improves this alignment in Qwen, we examine the modality gap between image and caption embeddings.

We randomly selected 35 samples from the MS-COCO test \ref{sec:coco} set and formed two groups:
\begin{itemize}
    \item Image with a blank white box (no text) and its caption generated with explicit textual instruction "Describe this image in detail".
    \item Image with an embedded question (Prompt-in-Image) and its caption generated without any textual instruction.
\end{itemize}
We fed both the image and its corresponding caption into the Qwen2.5-VL model and extracted final-layer embeddings. We then computed the average cosine distance between the image tokens and text tokens in the shared semantic space.
Results show that the Prompt-in-Image group consistently exhibits smaller cosine distances, with an average reduction of 12\%. In Figure~\ref{fig:token_embedding}, we also present PCA visualizations of two sets of test samples. The right plot (Prompt-in-Image) shows a smaller modality gap, where visual and text tokens are more closely aligned. This suggests that Prompt-in-Image acts as a bridge between the two modalities, effectively reducing the gap and enhancing multimodal alignment.

\section{Conclusion}
\label{sec:conclusion}
In this work, we propose Prompt-in-Image, a simple yet effective strategy that embeds textual instructions directly into images to unify the input modality. Through systematic evaluations on three representative VLMs, Qwen2.5-VL, InstructBLIP and LLaVA-1.5, we observe divergent effects: while Prompt-in-Image consistently improves Qwen’s performance and reduces hallucination, it significantly degrades InstructBLIP and LLaVA’s output quality.

We further analyze this phenomenon and identify key differences between the two models.On the one hand, InstructBLIP and LLaVA (based on CLIP) exhibit excessive attention to embedded text regions, leading to over-reliance on local patterns and results in hallucination. In contrast, Qwen demonstrates stronger robustness. On the other hand, Prompt-in-Image helps Qwen by enhancing modality fusion and mitigating alignment issues. Empirical results confirm that Prompt-in-Image leads to a smaller modality gap and improved cross-modal coherence.

This work shows that how models are trained on multimodal data really matters. It also suggests that simpler, unified approaches to VLM architecture might be worth exploring further.

\bibliography{anthology,custom}

\begin{thebibliography}{27}
\expandafter\ifx\csname natexlab\endcsname\relax\def\natexlab#1{#1}\fi

\bibitem[{Bai et~al.(2025)Bai, Chen, Liu, Wang, Ge, Song, Dang, Wang, Wang, Tang et~al.}]{bai2025qwen2}
Shuai Bai, Keqin Chen, Xuejing Liu, Jialin Wang, Wenbin Ge, Sibo Song, Kai Dang, Peng Wang, Shijie Wang, Jun Tang, et~al. 2025.
\newblock Qwen2. 5-vl technical report.
\newblock \emph{arXiv preprint arXiv:2502.13923}.

\bibitem[{Cho et~al.(2022)Cho, Yoon, Kale, Dernoncourt, Bui, and Bansal}]{cho2022fine}
Jaemin Cho, Seunghyun Yoon, Ajinkya Kale, Franck Dernoncourt, Trung Bui, and Mohit Bansal. 2022.
\newblock Fine-grained image captioning with clip reward.
\newblock \emph{arXiv preprint arXiv:2205.13115}.

\bibitem[{Dai et~al.(2023)Dai, Li, Li, Tiong, Zhao, Wang, Li, Fung, and Hoi}]{dai2023instructblipgeneralpurposevisionlanguagemodels}
Wenliang Dai, Junnan Li, Dongxu Li, Anthony Meng~Huat Tiong, Junqi Zhao, Weisheng Wang, Boyang Li, Pascale Fung, and Steven Hoi. 2023.
\newblock \href {http://arxiv.org/abs/2305.06500} {Instructblip: Towards general-purpose vision-language models with instruction tuning}.

\bibitem[{Darcet et~al.(2023)Darcet, Oquab, Mairal, and Bojanowski}]{darcet2023vision}
Timoth{\'e}e Darcet, Maxime Oquab, Julien Mairal, and Piotr Bojanowski. 2023.
\newblock Vision transformers need registers.
\newblock \emph{arXiv preprint arXiv:2309.16588}.

\bibitem[{Gong et~al.(2024)Gong, Ming, Wang, and Wei}]{gong2024damro}
Xuan Gong, Tianshi Ming, Xinpeng Wang, and Zhihua Wei. 2024.
\newblock Damro: Dive into the attention mechanism of lvlm to reduce object hallucination.
\newblock \emph{arXiv preprint arXiv:2410.04514}.

\bibitem[{Leng et~al.(2024)Leng, Zhang, Chen, Li, Lu, Miao, and Bing}]{leng2024mitigating}
Sicong Leng, Hang Zhang, Guanzheng Chen, Xin Li, Shijian Lu, Chunyan Miao, and Lidong Bing. 2024.
\newblock Mitigating object hallucinations in large vision-language models through visual contrastive decoding.
\newblock In \emph{Proceedings of the IEEE/CVF Conference on Computer Vision and Pattern Recognition}, pages 13872--13882.

\bibitem[{Li et~al.(2023)Li, Du, Zhou, Wang, Zhao, and Wen}]{li2023evaluating}
Yifan Li, Yifan Du, Kun Zhou, Jinpeng Wang, Wayne~Xin Zhao, and Ji-Rong Wen. 2023.
\newblock Evaluating object hallucination in large vision-language models.
\newblock \emph{arXiv preprint arXiv:2305.10355}.

\bibitem[{Li et~al.(2024)Li, Yang, Liu, Ma, Zhang, Yang, Sun, Liu, and Bai}]{li2024monkey}
Zhang Li, Biao Yang, Qiang Liu, Zhiyin Ma, Shuo Zhang, Jingxu Yang, Yabo Sun, Yuliang Liu, and Xiang Bai. 2024.
\newblock Monkey: Image resolution and text label are important things for large multi-modal models.
\newblock In \emph{proceedings of the IEEE/CVF conference on computer vision and pattern recognition}, pages 26763--26773.

\bibitem[{Liang et~al.(2022)Liang, Zhang, Kwon, Yeung, and Zou}]{liang2022mind}
Victor~Weixin Liang, Yuhui Zhang, Yongchan Kwon, Serena Yeung, and James~Y Zou. 2022.
\newblock Mind the gap: Understanding the modality gap in multi-modal contrastive representation learning.
\newblock \emph{Advances in Neural Information Processing Systems}, 35:17612--17625.

\bibitem[{Lin et~al.(2014)Lin, Maire, Belongie, Hays, Perona, Ramanan, Doll{\'a}r, and Zitnick}]{lin2014microsoft}
Tsung-Yi Lin, Michael Maire, Serge Belongie, James Hays, Pietro Perona, Deva Ramanan, Piotr Doll{\'a}r, and C~Lawrence Zitnick. 2014.
\newblock Microsoft coco: Common objects in context.
\newblock In \emph{Computer vision--ECCV 2014: 13th European conference, zurich, Switzerland, September 6-12, 2014, proceedings, part v 13}, pages 740--755. Springer.

\bibitem[{Liu et~al.(2023{\natexlab{a}})Liu, Lin, Li, Wang, Yacoob, and Wang}]{liu2023mitigating}
Fuxiao Liu, Kevin Lin, Linjie Li, Jianfeng Wang, Yaser Yacoob, and Lijuan Wang. 2023{\natexlab{a}}.
\newblock Mitigating hallucination in large multi-modal models via robust instruction tuning.
\newblock \emph{arXiv preprint arXiv:2306.14565}.

\bibitem[{Liu et~al.(2024)Liu, Xue, Chen, Chen, Zhao, Wang, Hou, Li, and Peng}]{liu2024survey}
Hanchao Liu, Wenyuan Xue, Yifei Chen, Dapeng Chen, Xiutian Zhao, Ke~Wang, Liping Hou, Rongjun Li, and Wei Peng. 2024.
\newblock A survey on hallucination in large vision-language models.
\newblock \emph{arXiv preprint arXiv:2402.00253}.

\bibitem[{Liu et~al.(2023{\natexlab{b}})Liu, Li, Wu, and Lee}]{liu2023visual}
Haotian Liu, Chunyuan Li, Qingyang Wu, and Yong~Jae Lee. 2023{\natexlab{b}}.
\newblock Visual instruction tuning.
\newblock \emph{Advances in neural information processing systems}, 36:34892--34916.

\bibitem[{Lovenia et~al.(2023)Lovenia, Dai, Cahyawijaya, Ji, and Fung}]{lovenia2023negative}
Holy Lovenia, Wenliang Dai, Samuel Cahyawijaya, Ziwei Ji, and Pascale Fung. 2023.
\newblock Negative object presence evaluation (nope) to measure object hallucination in vision-language models.
\newblock \emph{arXiv preprint arXiv:2310.05338}.

\bibitem[{Niu et~al.(2021)Niu, Tang, Zhang, Lu, Hua, and Wen}]{niu2021counterfactual}
Yulei Niu, Kaihua Tang, Hanwang Zhang, Zhiwu Lu, Xian-Sheng Hua, and Ji-Rong Wen. 2021.
\newblock Counterfactual vqa: A cause-effect look at language bias.
\newblock In \emph{Proceedings of the IEEE/CVF conference on computer vision and pattern recognition}, pages 12700--12710.

\bibitem[{OpenAI(2024)}]{openai2024gpt4o}
OpenAI. 2024.
\newblock Gpt-4o: Openai’s new flagship model.
\newblock \url{https://openai.com/index/gpt-4o}.
\newblock Accessed: 2025-05-23.

\bibitem[{Rabinovich et~al.(2023)Rabinovich, Ackerman, Raz, Farchi, and Anaby-Tavor}]{rabinovich2023predicting}
Ella Rabinovich, Samuel Ackerman, Orna Raz, Eitan Farchi, and Ateret Anaby-Tavor. 2023.
\newblock Predicting question-answering performance of large language models through semantic consistency.
\newblock \emph{arXiv preprint arXiv:2311.01152}.

\bibitem[{Radford et~al.(2021)Radford, Kim, Hallacy, Ramesh, Goh, Agarwal, Sastry, Askell, Mishkin, Clark et~al.}]{radford2021learning}
Alec Radford, Jong~Wook Kim, Chris Hallacy, Aditya Ramesh, Gabriel Goh, Sandhini Agarwal, Girish Sastry, Amanda Askell, Pamela Mishkin, Jack Clark, et~al. 2021.
\newblock Learning transferable visual models from natural language supervision.
\newblock In \emph{International conference on machine learning}, pages 8748--8763. PmLR.

\bibitem[{Rohrbach et~al.(2018)Rohrbach, Hendricks, Burns, Darrell, and Saenko}]{rohrbach2018object}
Anna Rohrbach, Lisa~Anne Hendricks, Kaylee Burns, Trevor Darrell, and Kate Saenko. 2018.
\newblock Object hallucination in image captioning.
\newblock \emph{arXiv preprint arXiv:1809.02156}.

\bibitem[{Role et~al.(2025)Role, Meyer, and Amblard}]{role2025fill}
Fran{\c{c}}ois Role, S{\'e}bastien Meyer, and Victor Amblard. 2025.
\newblock Fill the gap: Quantifying and reducing the modality gap in image-text representation learning.
\newblock \emph{arXiv preprint arXiv:2505.03703}.

\bibitem[{Sun et~al.(2023)Sun, Shen, Cao, Liu, Li, Shen, Gan, Gui, Wang, Yang et~al.}]{sun2023aligning}
Zhiqing Sun, Sheng Shen, Shengcao Cao, Haotian Liu, Chunyuan Li, Yikang Shen, Chuang Gan, Liang-Yan Gui, Yu-Xiong Wang, Yiming Yang, et~al. 2023.
\newblock Aligning large multimodal models with factually augmented rlhf.
\newblock \emph{arXiv preprint arXiv:2309.14525}.

\bibitem[{Wang et~al.(2024{\natexlab{a}})Wang, Zhou, Huang, Xu, Zhang, Poon, and Chen}]{wang2024mdpo}
Fei Wang, Wenxuan Zhou, James~Y Huang, Nan Xu, Sheng Zhang, Hoifung Poon, and Muhao Chen. 2024{\natexlab{a}}.
\newblock mdpo: Conditional preference optimization for multimodal large language models.
\newblock \emph{arXiv preprint arXiv:2406.11839}.

\bibitem[{Wang et~al.(2023)Wang, Wang, Xu, Zhang, Gu, Jia, Wang, Xu, Yan, Zhang et~al.}]{wang2023amber}
Junyang Wang, Yuhang Wang, Guohai Xu, Jing Zhang, Yukai Gu, Haitao Jia, Jiaqi Wang, Haiyang Xu, Ming Yan, Ji~Zhang, et~al. 2023.
\newblock Amber: An llm-free multi-dimensional benchmark for mllms hallucination evaluation.
\newblock \emph{arXiv preprint arXiv:2311.07397}.

\bibitem[{Wang et~al.(2024{\natexlab{b}})Wang, Pan, Ding, and Biemann}]{wang2024mitigating}
Xintong Wang, Jingheng Pan, Liang Ding, and Chris Biemann. 2024{\natexlab{b}}.
\newblock Mitigating hallucinations in large vision-language models with instruction contrastive decoding.
\newblock \emph{arXiv preprint arXiv:2403.18715}.

\bibitem[{Wang et~al.(2025)Wang, Hooi, Wang, Yang, Huang, and Cai}]{wang2025text}
Zhaochen Wang, Bryan Hooi, Yiwei Wang, Ming-Hsuan Yang, Zi~Huang, and Yujun Cai. 2025.
\newblock Text speaks louder than vision: Ascii art reveals textual biases in vision-language models.
\newblock \emph{arXiv preprint arXiv:2504.01589}.

\bibitem[{Yin et~al.(2025)Yin, Si, and Wang}]{yin2025mirage}
Hao Yin, Gunagzong Si, and Zilei Wang. 2025.
\newblock The mirage of performance gains: Why contrastive decoding fails to address multimodal hallucination.
\newblock \emph{arXiv preprint arXiv:2504.10020}.

\bibitem[{Zhang et~al.(2024)Zhang, Xie, Chen, Sun, Wang et~al.}]{zhang2024dhcp}
Yudong Zhang, Ruobing Xie, Jiansheng Chen, Xingwu Sun, Yu~Wang, et~al. 2024.
\newblock Dhcp: Detecting hallucinations by cross-modal attention pattern in large vision-language models.
\newblock \emph{arXiv preprint arXiv:2411.18659}.

\end{thebibliography}
\bibliographystyle{acl_natbib}

\appendix


\end{document}